\documentclass[10pt,twocolumn,letterpaper]{article}

\usepackage{cvpr}              

\def\sup3d{S-Up3D}
\def\nrsfs{ALLRAP}
\usepackage[dvipsnames]{xcolor}
\usepackage{enumerate}%

\definecolor{cvprblue}{rgb}{0.21,0.49,0.74}
\usepackage[pagebackref,breaklinks,colorlinks,allcolors=cvprblue]{hyperref}

\def\paperID{15} 
\def\confName{CVPR}
\def\confYear{2025}

\title{
Unsupervised 2D-3D lifting of non-rigid objects using local constraints
}

\author{
Shalini Maiti\textsuperscript{1,2}\quad 
Lourdes Agapito\textsuperscript{2}\quad 
Benjamin Graham\footnotemark \quad\\
\vspace{0.05cm} \\
\textsuperscript{1}Meta AI\quad
\textsuperscript{2}University College London
}

\begin{document}
\twocolumn[{%
\renewcommand\twocolumn[1][]{#1}%
\maketitle

\begin{center}
\centering
\captionsetup{type=figure}
\includegraphics [width=0.93\linewidth]{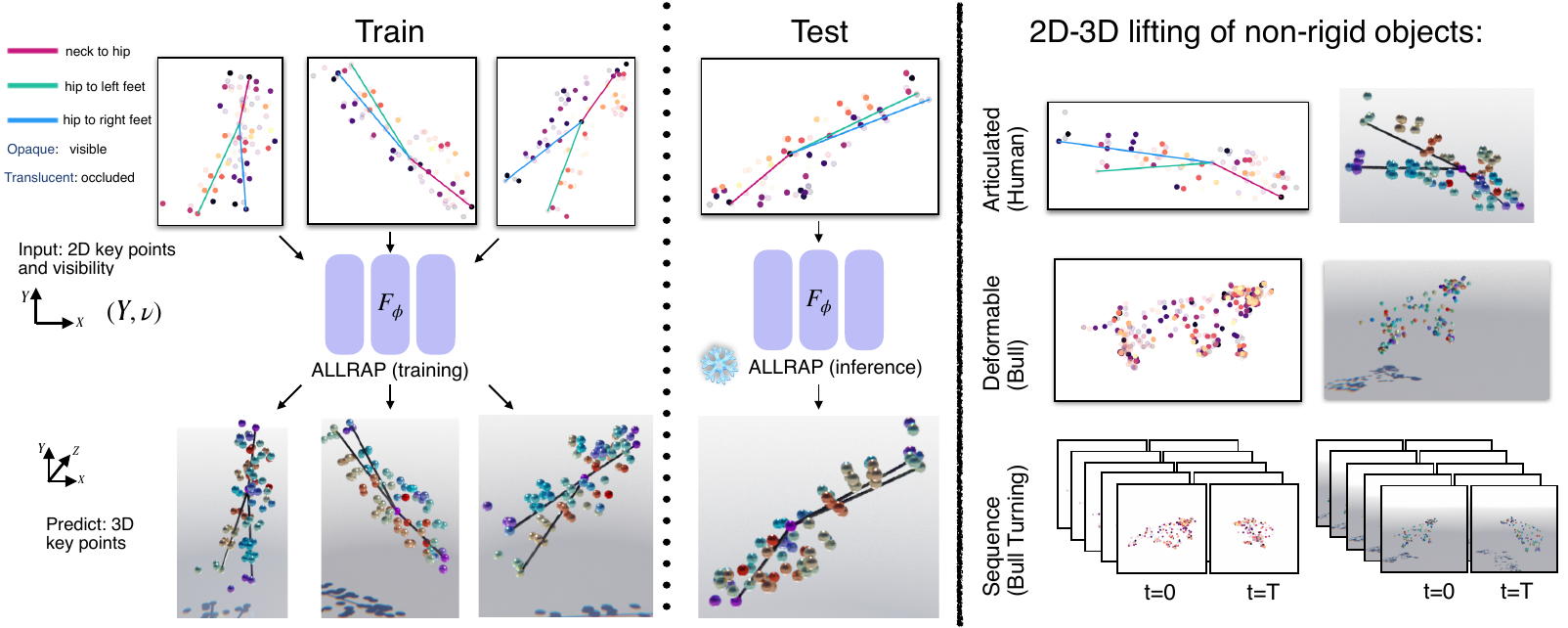}
\vspace{-2mm}
\caption{\textbf{3D Reconstruction for partially-occluded 2D semantic keypoints.} With an orthographic camera model, for visible points ($v=1$) we know the screen coordinates $(x,y)$ and must predict only the depth $z$. For occluded points ($v=0$), we must predict all three $(x,y,z)$-coordinates. At test time, a trained deep network predicts the unknown coordinates. The network is generic, without any kind of low-rank constraints built-in; the geometric intuition is built into the losses during training, see Figure~\ref{fig:splash}. For a perspective camera model, the same process can be applied along the camera rays, rather than directly using the $x$-, $y$- and $z$-axes.}
\vspace{-2mm}
\label{fig:teaser}
\end{center}%
}]

\renewcommand{\thefootnote}{$*$}
\footnotetext{Work done while working at Meta.}
\renewcommand{\thefootnote}{\arabic{footnote}}
\begin{abstract}
For non-rigid objects, predicting the 3D shape from 2D keypoint observations is ill-posed due to occlusions, and the need to disentangle changes in viewpoint and changes in shape. This challenge has often been addressed by embedding low-rank constraints into specialized models. These models can be hard to train, as they depend on finding a canonical way of aligning observations, before they can learn detailed geometry. These constraints have limited the reconstruction quality. We show that generic, high capacity models, trained with an unsupervised loss, allow for more accurate predicted shapes. In particular, applying low-rank constraints to localized subsets of the full shape allows the high capacity to be suitably constrained. We reduce the state-of-the-art reconstruction error on the \sup3d dataset by over 70\%. 

\end{abstract}    
\section{Introduction}
\label{sec:intro}
Unsupervised 2D-3D lifting of non-rigid objects is the problem of inferring the 3D shape of a deformable object such as animals solely from 2D observations. The solution to this open-ended problem could assist many applications in human-computer interaction, virtual/augmented reality and creative media generation. 

Early NRSfM techniques often made unrealistic assumptions about the data, for example, that they were restricted to sets of keypoints without occlusions, or to small numbers of keypoints. They were also often forced to use orthographic or weak perspective camera models because of the non-linearity introduced by perspective effects. 

NRSfM is a fundamentally ill-posed problem and therefore requires the use of various mathematical priors to avoid simply predicting uniform depth. One such prior is the low-rank shape prior \cite{BreglerHB00}, which assumes that the various shapes can be described using linear combinations of a small set of basis shapes. 
For temporal sequences, a popular prior is minimizing the change in shape between consecutive frames \cite{rethinkingNrsfm08,nrsfmKumar20}. 
\begin{figure*}[!ht]
\begin{center}
\includegraphics[width=\textwidth]{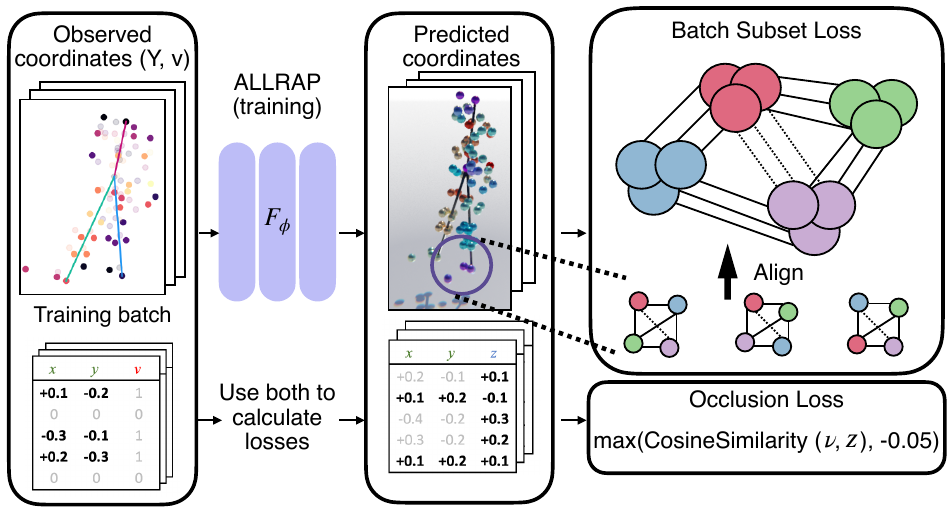}
\end{center}
\vspace{-0.2cm}
\caption{
\textbf{Unsupervised losses} Starting from a randomly-initialized generic deep network, we iteratively learn from batches of partially-occluded 2D-annotated training samples. Our training use two unsupervised, batch-wise losses.
The subset loss, see Section \ref{sec:tuple_loss}, acts on the batch of noisy 3D reconstructions. It selects a subset of nearby keypoints, aligns the sub-shapes by rotation and translation, and finally measures the size of the non-rigid motion using the log-product of the  singular value decomposition of the residual error matrix, i.e. the log Gramian determinant. This encourages the model to predict body parts that are as consistent as possible. The occlusion loss, see Section \ref{sec:occlusion},  encourages a weak negative correlation between the binary keypoint-visibility annotations and the predicted depths. This uses the fact that visible keypoints often hide other keypoints because they are closer to the camera.}
\label{fig:splash}
\vspace{-0.2cm}
\end{figure*}
DeepNRSFM \cite{DeepNRSfMPP} uses block sparsity to restrict the set of linear combinations to lie in a union of low-rank subspaces. 
C3DPO \cite{c3dpo} assumes that each shape can be associated with a canonical pose, with a low-rank shape basis constraining the range of motion within the canonical frame of reference. A network to predict the canonical pose has to be learnt alongside the shape basis regression network.

We take a different approach, using a deep network  that lifts 2D observations into the 3D space directly, by `inpainting' information missing due to occlusions and the camera projection. In particular, we find that parameter efficient MLP-Mixer models \cite{mlpMixer,mlpMixer2} can be trained to generalize well to new observations.

To train these models, we introduce a novel unsupervised loss that minimizes the variation in shape of local neighborhoods of the object, after allowing for rotation and translation.
This is suitable for objects tracking at least a semi-dense set of keypoints, e.g. around 50-100 keypoints per object.

To resolve flip ambiguity in the depth direction, we also use occlusion as an additional learning signal. Due to the nature of self-occlusion, being visible tends to be weakly negatively correlated with depth; points nearer to the camera tend to occlude points that are further away.

We apply our method both to isolated observations from a dataset of semantic keypoints, and to videos where arbitrary keypoints are tracked over the length of the video. We find that our method can be applied 
robustly across a range of different types of datasets, shapes and camera settings. 

\noindent In summary, our contributions are:
\begin{itemize}
\item We introduce a novel combination of unsupervised loss functions for the 2D-3D lifting of deformable shapes. 
\item We find that MLP-Mixers \cite{mlpMixer,mlpMixer2} are a parameter efficient alternative to the large MLPs that have been used in previous work on 2D-3D lifting using neural networks.
\item We achieve state of the art result on the \sup3d dataset, reducing the reconstruction error by over 70\% compared to prior works \cite{c3dpo,paul,DeepNRSfMPP,e2eMultiCat22}.
\item We show that our method can be also be applied in a one-shot fashion to lift arbitrary videos of tracked keypoints into 3D, with superior performance to baselines \cite{c3dpo,paul,DeepNRSfMPP}. 
\end{itemize}

\section{Related Work}
This section reviews the approaches for the 2D-to-3D lifting problem including classical baselines, parts based methods, and deep-learning based methods trained using unsupervised learning.

\textbf{Classical NRSfM} Tomasi and Kanade's \cite{TomasiKanade} approached the problem by fixing the rank of the rigid motion as three. Bregler et al. \cite{BreglerHB00} built on this low-rank assumption and formulated the problem of NRSfM as a linear combination of shape bases. Other works \cite{NRSfM-traj-space08,traj-space-dual-rep11,simplePriorFree12} have leveraged the low-rank prior, inspired from Bregler's factorization. Kumar \cite{nrsfmKumar20} argues that properly utilizing smoothness priors can result in competitive performance for video sequences. Low-rank constraints have informed many of the recent works applying deep-learning to NRSfM.

\textbf{Piecewise non-rigid 2D-3D lifting.}
The idea of piecewise 3D reconstruction has been explored before. \cite{Lee_2016_CVPR} and its extension \cite{Cha_2019_IEEE} treat 3D reconstruction as a consensus sampling problem, by first reconstructing subparts of the object, generating "weaker" hypotheses followed by optimisation over them to generate the final 3D reconstruction. \cite{Fayad_2010_ECCV} is another classical method that divides the surface into overlapping patches, individually reconstructs each patch using a  quadratic deformation model and registers them globally by imposing the constraint that points shared by patches must correspond to the same 3D points in space. 

Our training loss relies on the same intuition as the methods above, that  parts of an object can be simpler to model than the whole. However, rather than learning pieces and stitching them together later, our model learns to reconstruct the entire object in one forward pass, with piecewise constraints only being applied during training, to automatically selected subsets of the object.

\textbf{Deep non-rigid 2D-3D lifting.} Deep-learning methods \cite{deepNrsfm, c3dpo, DeepNRSfMPP, paul, e2eMultiCat22, deng2024deep,deepnrsfm_seq2seq,zeng2022mhr,wang2023temporal} have shown that 3D priors can be learnt from annotated 2D keypoints. 
DeepNRSfM \cite{deepNrsfm} solves NRSfM as a hierarchical block-sparse dictionary learning problem by utilising a deep neural network. However, their method is limited to weakly perspective and orthographic datasets with fully-annotated observations, which are not practically obtained in the real world settings. DeepNRSfM++ 
\cite{DeepNRSfMPP} build on top of this theoretical framework by adaptively normalising the 2D keypoints and the shape dictionary. They achieved competitive results in the cases of perspective datasets and self-occluded points. 
C3DPO \cite{c3dpo} jointly learnt viewpoint parameters and 3D shape of an object in an unsupervised manner using two deep networks for factorization and canonicalization. \cite{e2eMultiCat22} modifies C3DPO by imposing sparsity in the shape basis using a learnt threshold. Paul \cite{paul} showed that an auto-encoder with a low-dimensional bottleneck can be used to regularise 2D-to-3D lifting, along with a related 3D-to-3D autoencoder to encourage a form of pose canonicalization.

The deep learning methods mentioned above use complex architectures to enforce various types of low-rank constraints. In contrast, we will use a simple deep learning model, moving all the necessary regularization into the loss functions used for training. Whereas \cite{c3dpo,paul,deepNrsfm,DeepNRSfMPP,deepnrsfm_seq2seq,deng2024deep, zeng2022mhr} apply a low-rank/block-sparsity constraint to the entire shape, we will only encourage low-rankness in neighborhoods of the reconstruction. We believe this makes more sense as body parts like knees and elbows do exhibit very low-rank behaviour, while all the joints put together combine to make a body with a much higher number of degrees of freedom. Moreover, in case of sequences, we input the sequence as a batch, thereby learning temporal prior independent of the shape using our batched losses, making sequence reconstruction possible without the need of specialised sequence reconstruction modules~\cite{deepnrsfm_seq2seq,deng2024deep}.




\section{\nrsfs: As Locally Low-Rank As Possible}

We will set out our method in the orthographic case where the notation is slightly simpler, as camera rays are parallel. However, note that it can also be applied with a perspective camera model. The $4\times4$ perspective projection matrix maps diverging camera rays into parallel rays in normalized device coordinate (NDC) space. 

Let $(\mathbf{W},\mathbf{V})\in\mathbb{R}^{K,3}$ denote a collection of 2D keypoints and corresponding visibility masks. For visible keypoints $w_i=(x_i,y_i)$ is observed and $v_i=1$.
For occluded points $w_i$ is undefined (i.e. set to zero) and $v_i=0$. The  goal is to predict the 3D shape $X\in\mathbb{R}^{K,3}$ where $X_i= (x_i,y_i,z_i)$, consists of the two screen coordinates $(x_i,y_i)$ and  depth $z_i$.

\subsection{A Matrix Inpainting Approach}
We view the 2D-to-3D lifting problem as a matrix inpainting problem. We must fill in the $x$ and $y$ values for occluded points, and the $z$ values for all points. See Figure \ref{fig:teaser}.

Using only the 2D keypoints, we train a deep network $\phi:(\mathbf{W},\mathbf{V})\to\mathbf{X}$ to predict the missing values. We train $\phi$ using only batches of partially visible keypoints. Although we say $\phi$ predicts values in $\mathbf{X}$, we will only ever use the parts of the outputs that are not already available in the input. As a result, the final output can never collapse to a trivial solution, even if unsupervised losses tend to prefer solutions that are as simple as possible, such as collapsing the output to a single point.

Unlike \cite{c3dpo,paul}, we do not have a reprojection loss as the reprojection error is equal to zero by construction.

\subsection{Network architectures}
We consider two candidate architectures for $\phi$. The first is a simple MLP network, which maps the $3K$ inputs to $3K$ outputs via a number of hidden layers of arbitrary size. This is similar to the network architecture used in \cite{c3dpo}.

The second architecture is the MLP-Mixer \cite{mlpMixer,mlpMixer2}, which can be thought of as a simplified transformer \cite{vaswani2017attention}. The input layer splits the input into $K$ tokens of size 3, one per keypoint, and each token is mapped by a shared linear layer to a latent space of size $n$. The resulting $nK$ hidden units are then operated on by a sequence of transformer-like two-layer MLPs. Unlike transformers, there is no attention. Instead, the MLPs alternate between operating on $K$ tokens of size $n$, and by taking transposes of the hidden state operating on $n$ tokens of size $K$. This repeated transposing allows information to spread across the tokens and within the token latent space. 

\subsection{Unsupervised Losses}
We have two unsupervised losses. They both operate on batches of $B$ observations. For one of the losses in particular, we must have $B>1$, so \nrsfs\  must be trained using batch gradient descent, rather than pure stochastic gradient descent. Applying $\phi$ (and inpainting) to a batch of $B$ observations produces an output 3D reconstruction tensor of size $(B,K,3)$.

\subsubsection{The subset loss\label{sec:tuple_loss}}
The subset loss operates on slices of the full output tensor. Given a subset of the $K$ keypoints, i.e. $\{i_1,\dots,i_k\}\subset\{1,\dots,K\}$, we can extract a sub-tensor of size $(B,k,3)$. This contains the 3D coordinates for $k$ of the $K$ keypoints, for each sample in the batch.

\paragraph{Subset selection.} We consider two methods for picking subsets of keypoints. The first is simply random sampling, without replacement. If there are around 80 keypoints, we might pick a subset of size 16 or 32.

The second strategy is sampling keypoint neighborhoods.
This is a form of bootstrapping: we use the network output to pick neighborhoods, to further train the network. We reshape the batch output tensor to have shape $(K,3B)$. We then pick a random keypoint, and add on its $k-1$ nearest neighbors in $\mathbb{R}^{3B}$. 
As training progresses, the network output will hopefully converge to the real 3D shapes, and then the neighborhoods constructed in this way will increasingly correspond to semantic 3D regions of the object.
In the case of animals, a neighborhood might correspond to a body part like a knee or a shoulder. In those cases, we might expect the subset of points to have a simpler `low-rank' representation than the whole shape, e.g. knees and hinges tend to have one major degree of freedom.

As we use randomness or the model output, neither of the strategies relies on having a predefined skeleton describing how the keypoints fit together. In practice, we create 10 subsets for each training batch, and average the resulting subset-losses.

\paragraph{Aligning the batch of subsets.}
The subset loss calculates the log of the volume of the residual shape differences after they have been aligned as close as possible to each other by a collection of rigid transformations. 

Our input subtensor $\mathbf{C}$ has shape $(B,k,3)$. First, we remove the translation component from the predicted shape by placing the center of each batch item at the origin and obtain $C^\mathrm{centered}$. We achieve this by subtracting the mean from each 3D location per sample \textit{i}, 
\begin{equation}\label{eq:3}
    \underbrace{\mathbf{C}^\mathrm{centered}_{i,:,:}}_{k\times 3} = \mathbf{C}_{i,:,:} - \frac{1}{k}\; \sum_{j=1}^k \mathbf{C}_{i,j,:}.
\end{equation}
Next, we reshape this into a tensor of shape $(3B,K)$ and compute the Tomasi-Kanade-style SVD factorization  \cite{TomasiKanade}:
\begin{equation}\label{eq:4}
\underbrace{\mathbf{C}^\mathrm{centered}}_{3B \times k} = \underbrace{\mathbf{U}}_{3B \times 3B} \times \underbrace{\mathbf{\Sigma}}_{3B\times k} \times \underbrace{\mathbf{V}^{\text{T}}}_{k \times k}
\end{equation}
Next, we obtain the batchwise mean-shape $\mu$ using the right eigenvectors corresponding to the 3 largest singular value components,
\begin{equation}\label{eq:5}
    \underbrace{\mu}_{k \times 3} = \mathbf{V}_{:,:3} \times \mathbf{\Sigma}_{:3, :3}
\end{equation}
Next, we use the left eigenvectors to compute a set of pseudo-rotation matrices $\hat{U}$ that should align $C^\mathrm{centered}$ with $\mu$. 
Let 
\begin{equation}\label{eq:6}
\underbrace{\hat{U}}_{B\times3\times3} \equiv \mathrm{reshape}\;( \mathbf{U}_{:,:3} )
\end{equation}
We don't need to apply an orthogonality constraint here because we only use these matrices to compute whether $\mu$ needs to be mirror-flip inverted to correct for the flip-ambiguity in its construction via SVD. We get this flip value of $\pm1$ by taking the sign of the sum of the determinants of the pseudo-rotation matrices $\hat{U}$:
\begin{equation}\label{eq:7}
    \mathrm{flip} = \mathrm{sign}\left(\sum^{B}_{b=1}\mathrm{det}(\underbrace{\hat{U}_{ b,:,:}}_{3\times3})\right)
\end{equation}
If flip is negative, the $k\times3$ slices of $\mathbf{C}^\mathrm{centered}$ can be aligned by rotation more closely with $-\mu$ than $\mu$. In that case, we just replace $\mu$ with $-\mu$.

We now align the elements $\mathbf{C}^\mathrm{centered}_{i,:,:}$ with $\mu$ using the Kabsch-Umeyama \cite{kabsch1976,Kabsch1978,Umeyama} algorithm,
\begin{equation}\label{eq:9}
    R_i = \mathrm{K.U.}(\mathbf{C}^\mathrm{centered}_{i,:,:}, \mu)\in \mathrm{SO}(3).
\end{equation}
Aligning the batch samples to $\mu$ with the $R_i$ and subtracting $\mu$ gives us a residual error $\mathbf{C}^\mathrm{residual}\in\mathbb{R}^{B,K,3}$ that cannot be explained by rigid motion. 
\begin{equation}
\mathbf{C}^\mathrm{residual}_{i,:,:} =\mathbf{C}^\mathrm{centered}_{i,:,:} \times R_i - \mu.
\end{equation}
To make our loss scale invariant, we divide this 
by either (i) the average depth value (for a perspective camera model) or (ii) the standard deviation of $\mathbf{C}^\mathrm{centered}$ (for orthographic cameras).
Call this scaled residual error $E$ and reshape it to have size $(B,K3)$. Leting $\sigma_1,...,\sigma_n$ denote the non-zero singular value of $E$, our loss is now 
\begin{equation}
\mathcal{L}_\mathrm{subset} = \sum_{i=1}^{n} \log (\sigma_i).
\end{equation}
When $E\cdot E^\intercal$ is full rank, the loss is $\tfrac12 \log \det (E \cdot E^\intercal)$. Then $\mathcal{L}_\mathrm{subset}$ can also be described as the log of the Gramian determinant of the residual error matrix $E$.

\subsubsection{Occlusion Loss\label{sec:occlusion}}
This is the second component of our loss. 
Although occluded screen coordinates represent missing information, the visibility indicator function can convey information about the depth coordinates $z$. Self-occlusions occur when a part of an object that is closer to the camera occludes another part of the object that is further away, e.g. someone's face will occlude the back of their head if it is closer to the camera. We would therefore expect the binary visibility variable to be negatively correlated with depth. 

Where an entire body part is occluded, e.g. if someone is standing behind a table so their legs are not visible, then $v=0$ for the front and back of their legs, so the depth there is independent of $v$. Overall, we should expect visibility to be only weakly negatively correlated with depth. Visible points can be far away from the camera, and points near to the camera can be occluded.

Let $(v_i)_{i=1}^{BK}$ denote the vector of visibility indicator variables in the batch; $v_i=1$ if the $i$-th keypoint is visible, and $v_i=0$ if it is occluded. Let $(z_i)$ denote the predicted depth values of the corresponding points.

Let $v^c$ and $z^c$ denote the vectors obtained from $v$ and $z$ by subtracting their respective mean values. We are expecting the cosine similarity between $v^c$ and $z^c$ to be weakly negative as visible points tend to be closer to the camera. We therefore clamp the cosine similarity at $-0.05$ to avoid biasing the shape of the model. Our loss is therefore:
\begin{equation}
    \mathcal{L}_\mathrm{occlusion}(v, z) = \max\left(\frac{v^c\cdot z^c}{\|v^c\|_2 \cdot \|z^c\|_2},-0.05\right).
\end{equation}
Once the cosine similarity is below $-0.05$, which empirically tends to happen after just a few training iterations, the subset loss tends to push the cosine similarity below -0.05. The occlusion loss has then served its purpose, and the gradient of the loss goes to zero, leaving the subset loss to be the main training signal.

\begin{table}[!t]
\begin{centering}
\begin{tabular}{lcc}
\toprule 
Network(depth,width) & Parameters (M) & MPJPE\tabularnewline
\midrule 
MLP(2,1024) & 2.59 & 0.0548\tabularnewline
MLP(4,1024) & 4.69 & 0.0534\tabularnewline
MLP(6,1024) & 6.78 & 0.0542\tabularnewline
\midrule 
MLP-Mixer(8,32) & 0.24 & 0.0314\tabularnewline
MLP-Mixer(16,32) & 0.48 & 0.0289\tabularnewline
MLP-Mixer(24,32) & 0.72 & 0.0271\tabularnewline
MLP-Mixer(32,32) & 0.95 & 0.0260\tabularnewline
\bottomrule
\end{tabular}
\par\end{centering}
\caption{\textbf{Network design} We conduct a validation experiment using a 50:50 split of the \sup3d training data to explore the effect of deep network design on the reconstruction accuracy. The subset loss is applied with randomly chosen subset of size 32. MLP-Mixer networks generalize much better than MLP networks, whilst using far fewer parameters.
\label{tbl:mlp_vs_mixer}}
\end{table}
\begin{table}[!t]
\begin{centering}
\begin{tabular}{cccc}
\toprule 
Random & NN & MPJPE & Time (ms)\tabularnewline
\midrule 
8 & $\cdot$ & 0.0656 & 4.1\tabularnewline
16 & $\cdot$ & 0.0399 & 5.7\tabularnewline
32 & $\cdot$ & {\bf 0.0260} & 9.7\tabularnewline
48 & $\cdot$ & 0.0284 & 13.6\tabularnewline
\midrule 
$\cdot$ & 8 & 0.0269 & 4.1\tabularnewline
$\cdot$ & 16 & 0.0218 & 5.8\tabularnewline
$\cdot$ & 32 & {\bf 0.0191} & 9.7\tabularnewline
$\cdot$ & 48 & 0.0223 & 13.6\tabularnewline
$\cdot$ & 16,32 & 0.0198 & 15.3\tabularnewline
\midrule 
32 & 8 & 0.0230 & 13.7\tabularnewline
32 & 16 & 0.0212 & 15.2\tabularnewline
32 & 32 & 0.0210 & 19.2\tabularnewline
32 & 16,32 & {\bf 0.0190} & 24.8\tabularnewline
\midrule 
\multicolumn{2}{c}{---All 79 points---} & 0.0308 & 3.6\tabularnewline
\bottomrule
\end{tabular}
\par\end{centering}
\caption{\textbf{Subset selection} We conduct a validation experiments using a 50:50 split of \sup3d training data to explore the effect of subset selection on the effectiveness of the subset loss. The fourth column records the time needed to evaluate the subset loss on a CPU; this is only relevant during training. The most useful training signal seems to come from selecting local neighborhoods of medium size. \label{tbl:tuple_selection}}
\end{table}
\begin{figure}[!t]
\begin{center}
\includegraphics[width=0.95\linewidth]{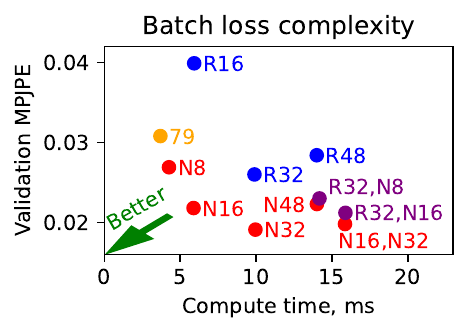}
\vspace{-6mm}
\end{center}
\caption{\textbf{Subset loss training-time efficiency} Plotting the results from Table~\ref{tbl:tuple_selection}, we see that selecting neigborhoods of 32 keypoints works well, whilst being reasonably fast at training time.}
\label{fig:tuple_selection}
\end{figure}

\section{Validation experiments}
We first ran a series of experiments training on 50\% of the \sup3d training dataset (see Section~\ref{sec:data}), validating on the rest of the training set, to set model- and loss-hyperparameters.
\subsection{Network architecture}
The first architecture we considered was a standard MLP, with 2, 4 or 6 hidden layers. We set the width to be 1024 units (c.f. \cite{c3dpo}) and put BatchNorm-ReLU activations after each hidden linear layer. 

The second architecture type is MLP-Mixer \cite{mlpMixer,mlpMixer2}, with ReLU activations and BatchNorm after the first linear module in each MLP and transposed-MLP. Each token has 32 units, and we tried 8,16,24 and 32 layers. Despite being much deeper, the MLP-Mixers have far fewer parameters, as they do not have any very large linear layers. 

All models are trained in the same way: using the subset loss with 10 randomly chosen subsets of size 32 per batch, and with the occlusion loss. The MLP-Mixers were trained with learning rate of $10^{-3}$, but the MLP networks had to be trained at a lower learning rate ($10^{-4}$) as otherwise they were unstable.

The results in Table~\ref{tbl:mlp_vs_mixer} show that the MLP-Mixer models reach much lower error rates than the MLP models. It seems the MLP-Mixer structure is particularly amenable to allowing keypoints to figure out their location relative to their neighbors, with their depth allowing a consensus shape to emerge among more distant body parts. We use an MLP-Mixer with depth and token-size 32 for the subsequent experiments.

\subsection{Subset selection}
To explore the subset loss, we trained identical MLP-Mixer networks, with depth and token-size 32, but using different strategies for picking the subsets, see Table~\ref{tbl:tuple_selection}. In Figure~\ref{fig:tuple_selection} we plot the reconstruction accuracy against the time it takes to calculate the subset loss. Up to experimental error, picking neighborhoods of size 32 is the simplest strategy that minimizes validation errors. We use this for the subsequent experiments.
\begin{table}[!t]
\begin{centering}
\begin{tabular}{lc}
\toprule 
Method & MPJPE\tabularnewline
\midrule 
EM-SfM \cite{EmSfm} & 0.107\tabularnewline
GbNRSfM \cite{NIPS2014_b53b3a3d} & 0.093\tabularnewline
DeepNRSfM \cite{DeepNRSfMPP} & 0.076\tabularnewline
C3DPO \cite{c3dpo} & 0.067\tabularnewline
Deep NRSfM++ \cite{DeepNRSfMPP}& 0.062\tabularnewline
PAUL \cite{paul} & 0.058\tabularnewline
e2eMulti \cite{e2eMultiCat22} & 0.057\\
Mini \nrsfs\ (ours) & 0.037\tabularnewline
\textbf{\nrsfs\ (ours)} & \textbf{0.0163}\tabularnewline
\bottomrule
\end{tabular}
\par\end{centering}
\caption{\textbf{Results on \sup3d} The results for \nrsfs\ is the average of 5 training runs with consecutive random seeds; the standard deviation across seeds is  0.0014.\label{tbl:up3d}}

\end{table}
\begin{figure*}[!ht]
\begin{center}
\includegraphics[width=\textwidth]{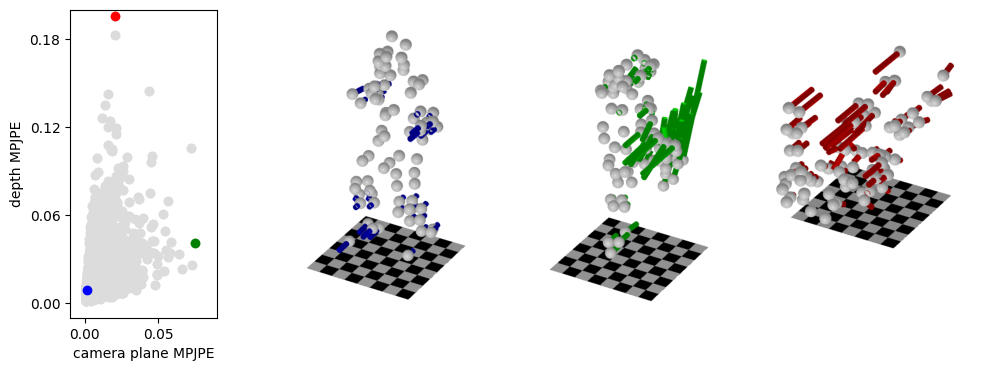}
\\
\hspace{1cm}(a)
\hspace{3.2cm}(b)
\hspace{4cm}(c)
\hspace{4cm}(d)
\end{center}
\vspace{-0.2cm}
\caption{\textbf{\nrsfs\ \sup3d test-set reconstructions.} (a) We break down the MPJPE error into errors in the camera plane due to occlusion, and errors in the predicted depths. (b) shows a test case with median errors in both components. (c) show the test cases with maximum errors in the camera plane; a leg is occluded and in an unusually high position. (d) shows the worst depth error: it is an unusual case as the body is observed from an over-the-head position. Errors are show in using, blue, green and red lines, respectively.
\label{fig:failure_cases}
}
\end{figure*}
\begin{figure*}[!ht]
\begin{center}
\includegraphics[trim={0cm 17cm 0cm 0cm},clip,width=\textwidth]{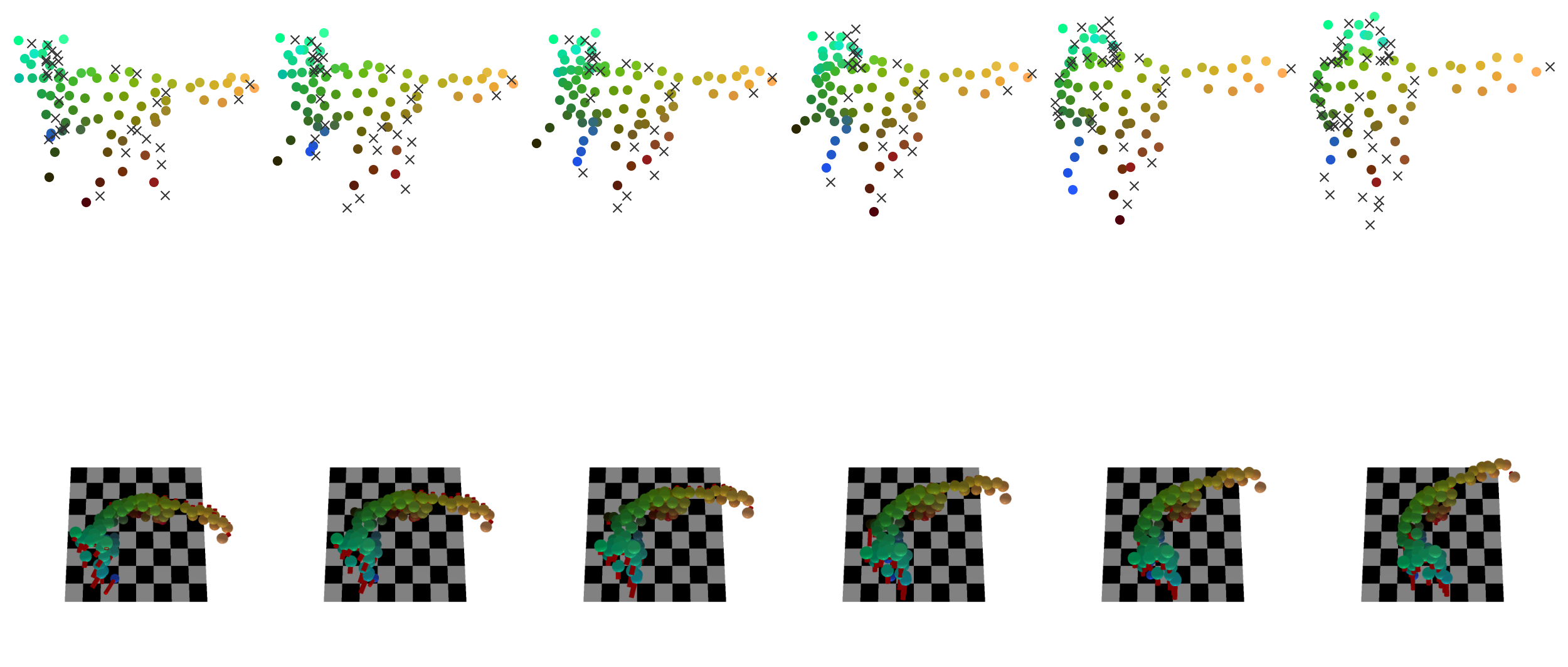}\\
\includegraphics[trim={0cm 0cm 0cm 16cm},clip,width=\textwidth]{figures/df4d_fox2}
\end{center}
\vspace{-0.2cm}
\caption{\textbf{\nrsfs\  DeformingThings4D one-shot reconstruction} Results from training \nrsfs\ on a sequence of 34 frames from the DeformingThings4D dataset. The reconstructions are shown from a top-down view, with errors shown with red lines. \label{fig:df4d}
}
\end{figure*}
\begin{figure*}[ht!]
\vspace{0cm}
\begin{center}
\includegraphics[trim={0cm 0cm 0cm 01cm},clip,width=1.0\textwidth]{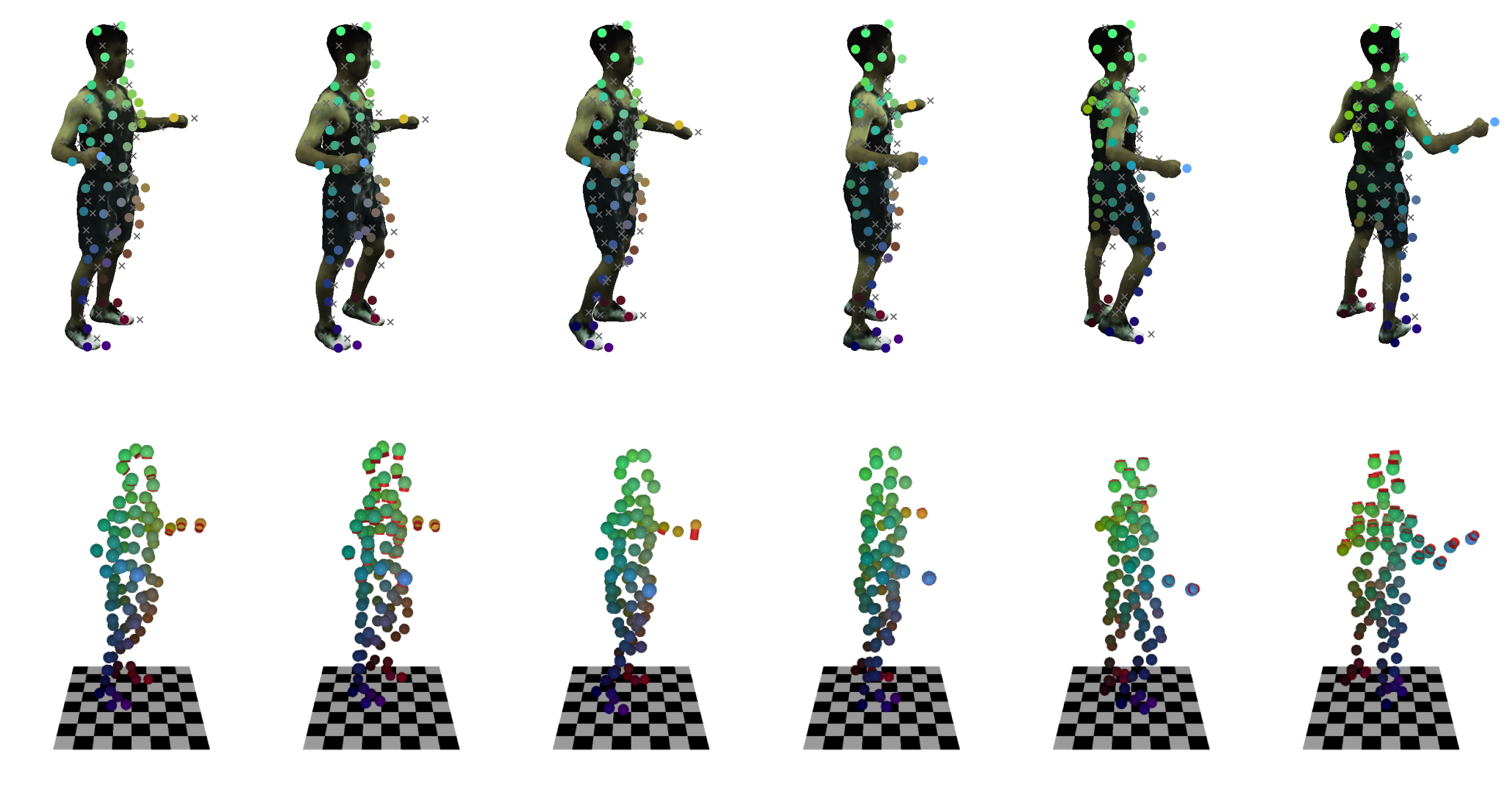}
\end{center}
\vspace{0cm}
\caption{{\textbf {\nrsfs\ ZJU-MoCap one-shot reconstructions}} Results from training \nrsfs\ on a single sequence of 570 frames from the ZJU-MoCap dataset annotated using \cite{FastCapture}. Errors compared to multicamera reconstruction are shown using red lines.  \label{fig:zju_mocap}}
\vspace{-0.2cm}
\end{figure*}
\begin{table*}[!ht]
\begin{centering}
{%
\begin{tabular}{lccccc}
\toprule 
Sequence & \#Frames & C3DPO\cite{c3dpo} & PAUL\cite{paul} & Deep NRSfM++ \cite{DeepNRSfMPP} & \bf{\nrsfs\ (ours)}\\ \midrule 
ZJU-377  & 570      &  0.372 & 0.1617 & 0.222  & {\bf 0.015}  \tabularnewline
ZJU-386  & 540      &  0.374 & 0.128 & 0.259  & {\bf 0.046}  \tabularnewline
ZJU-387  & 540      &  0.394 & 0.168 & 0.254  & {\bf 0.020}  \tabularnewline

\midrule
DF4D-Bull    & 26    &  0.456 & 0.272 & 0.325 & {\bf 0.167} \\
DF4D-Fox     & 34    &  0.359 & 0.117 & 0.280 & {\bf 0.073}  \\
DF4D-Puma    & 70    &  0.228 &  0.262 & 0.178 & {\bf 0.138} \\
DF4D-Lioness & 99    &  0.256 &  0.252 & 0.237 & {\bf 0.121} \\
\bottomrule
\end{tabular}
}
\par\end{centering}
\caption{\textbf{One-shot reconstruction results} We trained and evaluated \nrsfs  on single sequences from DeformingThings4D and ZJU-MOCAP/RTH. We compare with the three strong baseline methods from \sup3d \cite{c3dpo,paul,DeepNRSfMPP}. \label{tbl:deforming}}

\end{table*}






\section{Datasets\label{sec:data}}
We consider the following datasets of semi-dense 2D keypoints with 3D ground truths to test our method. They offer us a diverse sampling across object categories, camera projections, and deformations. 

\paragraph{Synthetic Up3D (\sup3d) \cite{Lassner:UP:2017}}
This is a dataset of human poses across different frames prepared as a synthetic 2D/3D projection of dense human keypoints based on the Unite the People 3D (Up3D) dataset. It does not contain any temporal information between observations. The camera projection is orthographic. The dataset is split into a training set of 171090 samples (inflated using data augmentation in the form of 2D rotation in the camera plane), and a test set of 15000 samples.

\paragraph{DeformingThings4D \cite{li20214dcomplete}} It is a collection of animated meshes corresponding to synthetic keypoints on animated 3D models of animals and people. We use 4 sequences containing a bull, a fox, a puma, and a lioness. To experiment with `low-shot' learning, we treat each video sequence of between 25 to 100 frames as a single dataset. We train and report reconstruction accuracy on the whole sequence. This tests the ability of methods to reconstruct novel objects from very limited observations.

The videos use a static perspective camera, so some points on the mesh are never visible, or only visible on the boundary of a small number of frames. We cannot expect to reconstruct these points, so we removed keypoints with less than 30\% visibility. We then downsampled the remaining keypoints to a set of 100 keypoints using farthest point sampling. 

\paragraph{ZJU-Mocap/RTH} The ZJU-Mocap dataset \cite{peng2021neural} is collection of synchronized multicamera videos. RealTimeHumans \cite{FastCapture} uses the multiple views to predict keypoints on a dense mesh, with the mesh deformed in 3D to match the multi-view photometric constraints. We project the keypoints into a single camera view, and train our model 
to reconstruct the 3D locations from the partially occluded 2D keypoints. Again, we treat each sequence of about 500 frames as a separate dataset to test 'one-shot' learning. We picked 100 keypoints per sequence using farthest point sampling.

\paragraph{Dataset Metrics}
The main metric used across the datasets is the Mean Per Joint Position Error, \begin{align}
\mathrm{MPJPE}(X^P, X^G) = \frac{1}{K} \sum_{k=1}^{K} \| X^P_k - X^G_k \|_2 
\end{align} where $X^P,X^G \in \mathbb{R}^{K,3}$ are the predicted and ground truth 3D shapes, respectively.

On \sup3d we calculate the MPJPE up to one degree of freedom, as the $z$ values can only be reconstructed up to a constant. Unlike \cite{c3dpo,paul,DeepNRSfMPP}, we don't calculate MPJPE twice, once with $z$ and once with $-z$, as thanks to the occlusion loss we don't suffer from mirror-flip ambiguity.

On ZJU-Mocap and DeformingThings4D there is scale ambiguity. We center each 3D reconstruction and ground truth shape. We then use regression to pick an optimal scale parameter to scale the predictions to match the ground truth. The same scale parameters is used over the whole temporal sequence.

\section{Experiments and results}
For each dataset, we trained a model with hyperparameters derived from the validation experiments. We trained an MLP-Mixer with depth 32, token-size 32, with the subset loss acting on neighborhoods of size 32. 

For \sup3d, we compare our results to existing methods in Table~\ref{tbl:up3d}. \nrsfs\ has a reconstruction error over 70\% lower than the next best existing methods. In Figure~\ref {fig:failure_cases} we break down the error in terms of camera-plane $(x,y)$ for occluded points, and depth errors. We show a typical 3D reconstruction and two failure cases. 

We also trained a very small MLP-Mixer on \sup3d,  with 8 layers and a token-size of 8; this is labelled Mini \nrsfs\ in Table~\ref{tbl:up3d}. Even this tiny network outperforms the prior methods.

In Table~\ref{tbl:deforming} we give results for single sequence `one-shot' reconstruction on the ZJU-Mocap dataset and the DeformingThings4D dataset. See Figure~\ref{fig:df4d} and Figure~\ref{fig:zju_mocap} for illustrations of the output.

It can be hard to see the details of the 3D reconstructions when reprojected back onto a 2D page. We include videos of the reconstructions in the supplementary material.

\section{Conclusion}

We have introduced a new method for non-rigid structure from motion. It exhibits strong performance on a variety of datasets compared to recent deep-learning based methods for 2D-to-3D lifting.
We will open source our implementation of the models and the training losses.

{
\small
\bibliographystyle{ieeenat_fullname}
\bibliography{main}
}

\end{document}


\title{NRSfS: Non-Rigid Structure from Subsets\\Supplementary}

\author{Shalini Maiti\\
Meta AI, University College of London\\
London\\
{\tt\small shalinimaiti@meta.com; shalini.maiti.21@ucl.ac.uk}
\and
Benjamin Graham\\
Meta AI\\
London \\
{\tt\small benjamingraham@meta.com}
}

\maketitle
\ificcvfinal\thispagestyle{empty}\fi

\section{Videos}
We include videos showing sequences from the ZJU-MoCap/FastCapture dataset and the DeformingThings4D dataset.
For ZJU-MoCap, we show the first 250 frames at 30 frames per second. For DeformingThings4D, we show all the frames at 5 frames per second.

\section{FastCapture}
FastCapture is concurrent work used to extract 2D keypoints from the ZJU-Mocap dataset. FastCapture is trained using photometric losses using multiple camera views. For training NRSfS, we only use the 2D keypoints locations visible from one of the cameras. We include the anonymised submission pdf.

\section{Training details}
\paragraph{S-Up3D.} The networks are 300 epochs with the Adam optimizer.
For MLP networks, the learning rate is $10^{-4}$, and for MLP-Mixer networks it is $10^{-3}$. Adaptive gradient clipping is used with a parameter of $0.1$.
\paragraph{ZJU-Mocap and DeformingThings4D.} These are trained for 200k iterations Adam, learning rate $10^{-5}$, adaptive gradient clipping 0.1.

\section{Typos}
We applogize for a small number of typos:
\begin{itemize}
\item L29, L145: 'Up3D' should read 'S-Up3D'.
\item L286: 'pick a random keypoints' should read 'pick a random keypoint'.
\item L304: 'as closes as possible' should  read 'as close as possible'.
\item L368: 'When $E$ is full rank' should read 'When $E\cdot E^\intercal$ is full rank.
\item L381-382: 'indicator varibles' should read 'indicator variables'.
\item L382: '$i$-th keypoints' should read '$i$-th keypoint'.
\item L418, L443, L462, L483: 'tuple' should read 'subset'.
\item L420: 'bu the MLP' should read 'but the MLP'.
\item L482: 'stratergies' should read 'strategies'.
\item L514-515: 'uses the multiple views' should read 'uses multiple views'.
\item L570: 'traing set' should read 'training set'.
\item L585: 'material, which' should read 'material.'
\end{itemize}